# Relevant Word Order Vectorization for Improved Natural Language Processing in Electronic Healthcare Records

Jeffrey Thompson[1,2], Jinxiang Hu[1,2], Dinesh Pal Mudaranthakam[1,2], David Streeter[1,2], Lisa Neums[1,2], Michele Park[2], Devin C. Koestler[1,2], Byron Gajewski[1,2], Matthew S. Mayo[1,2]

Affiliations:

1: Department of Biostatistics, University of Kansas Medical Center, Kansas City, KS, USA.

2: University of Kansas Cancer Center, Kansas City, KS, USA.

## Abstract

**Objective:** Electronic health records (EHR) represent a rich resource for conducting observational studies, supporting clinical trials, and more. However, much of the relevant information is stored in an unstructured format that makes it difficult to use. Natural language processing approaches that attempt to automatically classify the data depend on vectorization algorithms that impose structure on the text, but these algorithms were not designed for the unique characteristics of EHR. Here, we propose a new algorithm for structuring so-called free-text that may help researchers make better use of EHR. We call this method Relevant Word Order Vectorization (RWOV).

**Materials and Methods:** As a proof-of-concept, we attempted to classify the hormone receptor status of breast cancer patients treated at the University of Kansas Medical Center during a recent year, from the unstructured text of pathology reports. Our approach attempts to account for the semi-structured way that healthcare providers often enter information. We compared this approach to the ngrams and word2vec methods.

**Results:** Our approach resulted in the most consistently high accuracy, as measured by F1 score and area under the receiver operating characteristic curve (AUC).

**Discussion:** Our results suggest that methods of structuring free text that take into account its context may show better performance, and that our approach is promising.

**Conclusion:** By using a method that accounts for the fact that healthcare providers tend to use certain key words repetitively and that the order of these key words is important, we showed improved performance over methods that do not.

## Background and Significance

Since 2015, most healthcare providers in the United States have been required by the federal government to use electronic medical records (EMR), or suffer penalties to Medicaid and Medicare reimbursement levels [1]. Incentives were also provided to create and use electronic health records (EHR), which are more comprehensive, although the terms are frequently used interchangeably. In anticipation of that date, many healthcare providers transitioned to EHR well in advance. Consequently, many hospitals now have years of patient records stored in a relatively easy to search fashion. In the era of big data, researchers quickly saw the potential to utilize EHR to advance their work in ways that were previously challenging. There are many examples of how they might be used [2-6]. For example, with appropriate consideration of sample bias, retrospective cohorts for exposures captured in patient records might relatively easily be assembled to study the impact of things like smoking or BMI on outcomes for diseases that are commonly treated at hospitals, such as cancer [5]. EHR have also greatly simplified the process of recruiting subjects for prospective studies and clinical trials [6].

Likely, the biggest challenge in the use of EHR comes from extensive reliance on unstructured data. Healthcare providers typically use freeform notes to capture important information when interacting with patients. Frequently, these are simply typed into the EHR. Certain types of reports, such as pathology reports, are also entered as free text. Although efforts have been made to provide structure to the data when possible (e.g. checkboxes, numeric fields, or drop-down lists). It is not possible to reduce all patient interaction and information to a simple form. Often, some of the most important information is stored in these fields, such as patient descriptions of symptoms, or clinicians' observation of relevant signs of disease. Due to the volume of data generated by hospitals and other healthcare facilities, collecting data from free text is an arduous process to perform manually. Yet, particularly for clinical trials, this is currently the only available option at many institutions. Given the importance of clinical trials to drug development for a range of conditions, there is a critical need for methods that can facilitate this process. This led many researchers to propose the application of natural language processing (NLP) techniques to these data [7-12]. NLP is not a specific method but rather a collection of approaches that involve extracting information from language as it is naturally spoken or written. Increasingly, NLP efforts have been focused on EHR, to enable researchers to capitalize on its valuable information.

NLP is an extensive research area; thus, generalizations are difficult. However, many approaches are based on two main phases: data structuring and machine learning. In the data structuring phase, an algorithm attempts to impose some sort of regular structure on the data, in the machine learning phase, the newly structured data are used to learn some characteristic of the data (e.g. hormone receptor status for breast cancer patients). Often, supervised learning methods are used, therefore, these approaches require manually labeled data for each characteristic that one desires to be able to extract automatically. A challenge of the data structuring phase is to find some way to impose structure on data that allows meaningful information to be extracted. There is typically a great deal of information in the arrangement of natural language, but it also typically varies a great deal in its arrangement from instance to instance. Therefore, it is a challenge to reduce natural language to a structure that works well in most instances.

The field of NLP, as it is applied to EHR, is still developing, and there is a need for methods that are designed specifically to capitalize on the context of EHR. It is our hypothesis that methods specifically designed to exploit the nature of free text in the EHR should exhibit better performance than more general approaches. We note that while some fields may be unstructured, healthcare providers often state things in similar ways. Therefore, we expect a method that capitalizes on the repetitive nature of the text should show better performance than a more standard approach. Additionally, the order of text in a medical record is frequently important, as multiple results might be included near each other in the text (e.g. "positive for ER, negative for PR"). As part of our own work to support research at the University of Kansas Cancer Center using the EHR [13], we are investigating methods for using NLP to extract information from free text fields. In this article, we propose one possible such approach that is generalizable to other situations, and examine its performance in comparison to a couple of standard NLP methodologies.

## Methods

### Relevant Word Order Vectorization

In this work, we focus on the data structuring part of the NLP problem, thus we will pair our vectorization approach with a couple of different machine learning algorithms to compare performance to other methods. The basis of this method is to determine the structure of the most relevant words to predicting the class of text. Therefore, we call this approach Relevant Word Order Vectorization (RWOV). Pseudocode for the basic approach is shown in Algorithm 1.

*Algorithm 1*

```
for i from 1 to length of text blocks:
      block := text block i
      sentences := sentence tokenize block
      do:
            if toi in sentence:
                  toi_sentence := sentence
      until toi in sentence
      words := tokenize toi_sentence
      words := filter excluded words
      words := stem words
      patient[i] := words
      append words to all

top := most frequent n words from all
T := m x length of top matrix initiated to δ
for i from 1 to m:
      for j from 1 to length of top:
            s := if top[j] before toi in words then -1 else 1
            d := s x number of top words between top[j] and toi in patient[i]
            T[i,j] := if d = 0 then 0 else 1/d
```

The idea behind RWOV is quite simple. The approach is focused on predicting the class of a term of interest (TOI) from a block of text. Although EHR data are unstructured, nevertheless, there is a relatively concise vocabulary that is used by healthcare professionals when describing patient characteristics. Therefore, we should see the same terms occurring repeatedly in patient medical records. Furthermore, we propose that only a fraction of these terms indicate the meaning in relation to some particular term of interest. Nevertheless, we believe that the relative order of these most relevant words is important to the meaning of the text. RWOV creates a matrix, where each row represents a subject and each column a word. The words are those that co-occur the most frequently with some TOI. We will call these the top words. The value in each cell of the matrix is either 0, or the inverse of the number of top words that occur between the top word represented by the column and the TOI plus 1. The sign of the value indicates if the top word occurs before or after the TOI in the text. Cells are assigned a default value of 0. Therefore, the value in each cell drops away naturally in a nonlinear fashion from 1 (as close as possible to the TOI) to 0 (does not occur in the block of text).

## Data

For this study, we used a straightforward dataset to evaluate the performance of our NLP approach compared to a few other approaches. The dataset contains tumor pathology reports of women with breast cancer who sought treatment at the University of Kansas Medical Center in a recent year. Our goal is to identify the status of three important breast cancer biomarkers from the pathology report free text, biomarkers include estrogen receptor (ER), progesterone receptor (PR), and human epidermal growth factor receptor 2 (HER2). In order to keep the results more interpretable, we limited the datasets to include only those reports that included a determination of hormone receptor status. The number of positive and negative subjects for each hormone receptor are shown in Table 1.

*Table 1: Counts of subjects with hormone receptors status*

| Hormone Receptor | Positive (%) | Negative (%) | Total |
| --- | --- | --- | --- |
| Estrogen receptor (ER) | 227 (77.5) | 66 (22.5) | 293 |
| Progesterone receptor (PR) | 175 (63.9) | 99 (36.1) | 274 |
| Human epidermal growth factor 2 (HER2) | 14 (14.9) | 80 (85.1) | 94 |

## Analysis

Class imbalance is likely to be a factor when assessing performance with these types of data. Hormone receptor status is unbalanced in the population. Furthermore, depending on the study, researchers may be interested in one class or the other of the subjects. Complicating this, is the fact that some performance metrics, such as accuracy, can give the impression of good performance even when they are unable to accurately predict the class of interest. Therefore, we will break down the performance by class (and train separate models to predict each class), and provide F1 and AUC as our major performance metrics, given that accuracy can be misleading in these circumstances. We will not attempt to use sophisticated class balancing approaches in this assessment, so that we can minimize the number of factors that are being considered. F1 is defined as follows:

$$F1 = \frac{2 \times \frac{TP}{TP+FP} \times \frac{TP}{TP+FN}}{\frac{TP}{TP+FP} + \frac{TP}{TP+FN}}$$

where TP, FP, and FN stand for the number of true positives, false positives, and false negatives respectively. In other words, it is the harmonic mean of the precision and the recall. We will assume that each of our NLP methods can produce a score representing its confidence in the predicted class label of a sample. For this work, we will also assume that a sample can only be one of two classes. Then the AUC is simply the probability that a random sample that is truly of one class is scored higher than a random sample from the other class.

We will compare the performance of our approach to two popular vectorization methods. The first is known as n-grams [9 11], and the other is called word2vec [14], combined with either of two machine learning algorithms: support vector machines (SVM), and artificial neural networks (NN). For n-grams, we rely on the implementation in CountVectorizer module of the scikit learn library for python. For SVM, we used the SVC module, and for NN we use MLPClassifier. For word2vec, we used the genism library for python, and chose the skip-gram model. Our approach depends on two important stages, with their own associated algorithms: (1) data structuring or vectorization and (2) learning and prediction. At both of these stages, important choices can be made that affect performance. At the first stage, vectorization, there are hyperparameters for all of these approaches. For our approach, the only hyperparameter is the number of top words to model. This was decided by training and testing on an independent dataset and then this setting was simply used for all the analyses here. This will be the recommended default settings for our method, but likely performance could be improved by using training, validation, and test data. However, we wanted to preserve our sample size in this case. For n-grams, we used a number of different settings to try and determine the effect of considering greater or fewer numbers of words. These were [1,2], [2,2], [1,3], [2,3], and [3,3]. The numbers represent the ranges of the number of words to build n-grams from. Furthermore, the vectors were transformed by IDF [15]. The word2vec approach has more hyperparameters, which makes the choice more difficult. In this case, we searched groups of settings on an independent dataset and settled on some that are in a relatively common range (size = 200 for dimensionality of the vectors, window = 6 for the maximum distance between a word and the predicted word, negative = 5 for sampling negative words, and min_count = 3 for the minimum frequency of a word). The neural network structure was determined using a grid search to determine the best structure for the n-grams algorithm, and then using this structure for our own method, in order to give n-grams any possible advantage. Again, better performance could be achieved by tailoring this solution. As noted, we have taken measures to ensure the comparison is as fair as possible. In all comparisons, the exact same training and test data were used to compare all models. All results are presented as the average result over a threefold cross-validation.

## Results

Our results (Table 2) showed that RWOV has consistently high accuracy in detecting ER, PR, and Her2 status compared to other vectorization approaches, using either SVM or NN as a classifier. Furthermore, this result is true irrespective of which class is being predicted. In terms of class imbalance, the other approaches saw a noticeable decline in performance for the underrepresented class in most cases. This is despite the fact that class weighting was enabled for SVM. In every case, RWOV had the highest F1 score. This is particularly notable for the HER2+ class, which included only 14 subjects that were HER2+. It is worth noting that a method might achieve a much better AUC than F1, if the probability of being predicted as the opposite class was higher, because F1 takes the predicted class into account, and AUC only considers the relative probabilities of a single class.

*Table 2*

|  | ER+ | | ER- | | PR+ | | PR- | | HER2+ | | HER2- | |
|---|---|---|---|---|---|---|---|---|---|---|---|---|
|  | F1 | AUC | F1 | AUC | F1 | AUC | F1 | AUC | F1 | AUC | F1 | AUC |
| **RWOV-NN** | **0.96** | 0.95 | **0.87** | 0.95 | **0.96** | **0.98** | **0.93** | **0.98** | **0.71** | **0.96** | **0.94** | 0.9 |
| **RWOV-SVM** | **0.96** | 0.92 | 0.87 | 0.92 | 0.8 | 0.79 | 0.68 | 0.79 | 0.62 | 0.9 | 0.93 | **0.91** |
| **SVM(1,2)** | 0.93 | 0.95 | 0.74 | 0.95 | 0.90 | 0.94 | 0.81 | 0.94 | 0.22 | 0.53 | 0.92 | 0.68 |
| **SVM(2,2)** | 0.93 | **0.96** | 0.73 | **0.96** | 0.89 | 0.95 | 0.79 | 0.94 | 0.36 | 0.75 | **0.94** | 0.75 |
| **SVM(1,3)** | 0.93 | 0.95 | 0.71 | 0.95 | 0.89 | 0.94 | 0.8 | 0.94 | 0.24 | 0.72 | 0.92 | 0.72 |
| **SVM(2,3)** | 0.93 | 0.95 | 0.71 | 0.95 | 0.89 | 0.94 | 0.79 | 0.94 | 0.35 | 0.74 | **0.94** | 0.74 |
| **SVM(3,3)** | 0.92 | 0.94 | 0.67 | 0.95 | 0.88 | 0.93 | 0.79 | 0.93 | 0.35 | 0.75 | **0.94** | 0.74 |
| **SVM-W2V** | 0.76 | 0.67 | 0.42 | 0.67 | 0.70 | 0.71 | 0.55 | 0.73 | 0.16 | 0.60 | 0.69 | 0.52 |
| **NN(1,2)** | 0.93 | 0.9 | 0.77 | 0.92 | 0.89 | 0.92 | 0.76 | 0.92 | 0.24 | 0.48 | 0.92 | 0.49 |
| **NN(2,2)** | 0.94 | 0.91 | 0.73 | 0.91 | 0.89 | 0.91 | 0.78 | 0.92 | 0.25 | 0.63 | **0.94** | 0.68 |
| **NN(1,3)** | 0.92 | 0.86 | 0.7 | 0.93 | 0.88 | 0.93 | 0.75 | 0.91 | 0.25 | 0.61 | 0.93 | 0.48 |
| **NN(2,3)** | 0.93 | 0.83 | 0.68 | 0.9 | 0.88 | 0.9 | 0.72 | 0.89 | 0.24 | 0.55 | 0.93 | 0.54 |
| **NN(3,3)** | 0.92 | 0.81 | 0.63 | 0.9 | 0.87 | 0.89 | 0.67 | 0.87 | 0.13 | 0.49 | 0.93 | 0.56 |
| **NN-W2V** | 0.83 | 0.72 | 0.39 | 0.67 | 0.80 | 0.76 | 0.59 | 0.76 | 0 | 0.42 | 0.89 | 0.43 |
| **AIM** | 0.84 | | 0.5 | | 0.95 | | **0.93** | | 0.15 | | 0.76 | |

We created 95% bootstrap confidence intervals for all F1 scores and AUCs. These are shown in Figure 1 and Figure 2 respectively. ROC curves are shown in Figure 3.

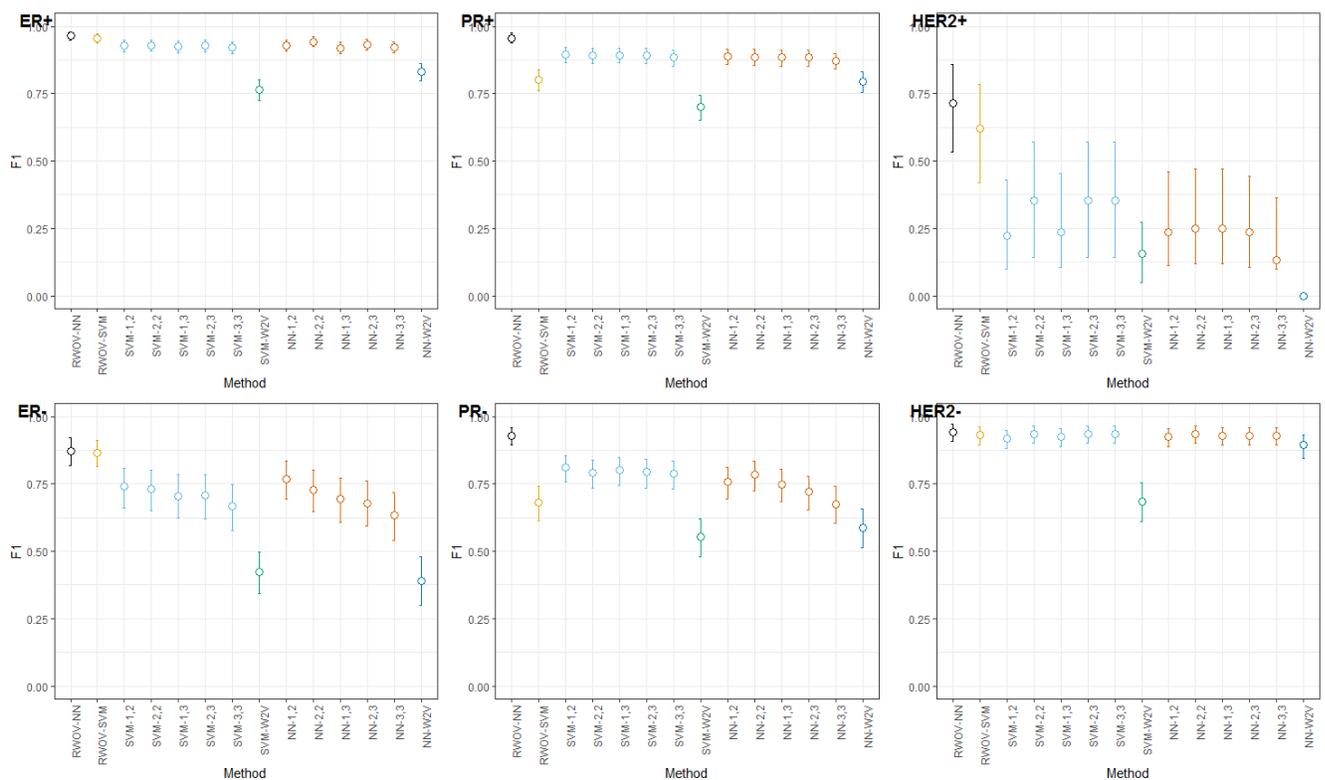

*Figure 1: 95% confidence intervals for F1 scores.*

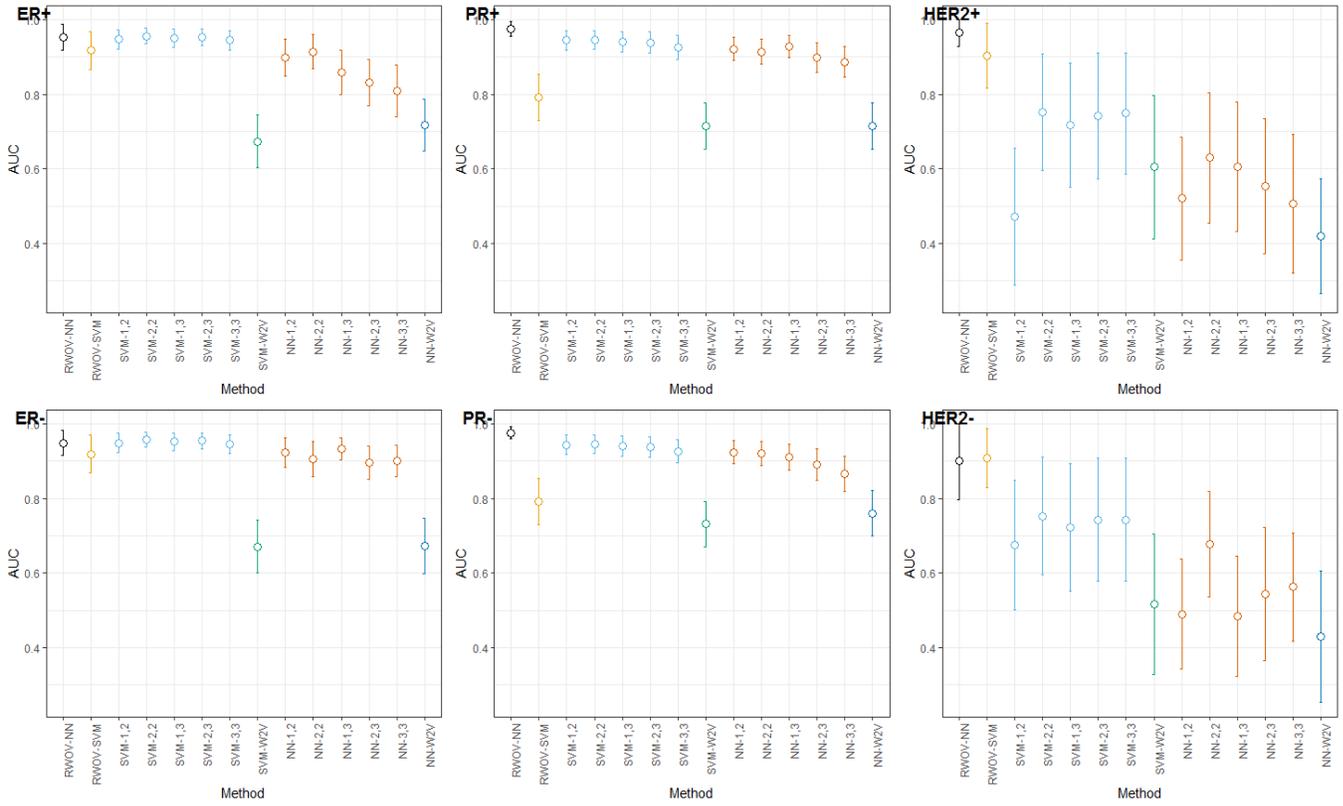

Figure 2: 95% confidence intervals for ROC AUCs.

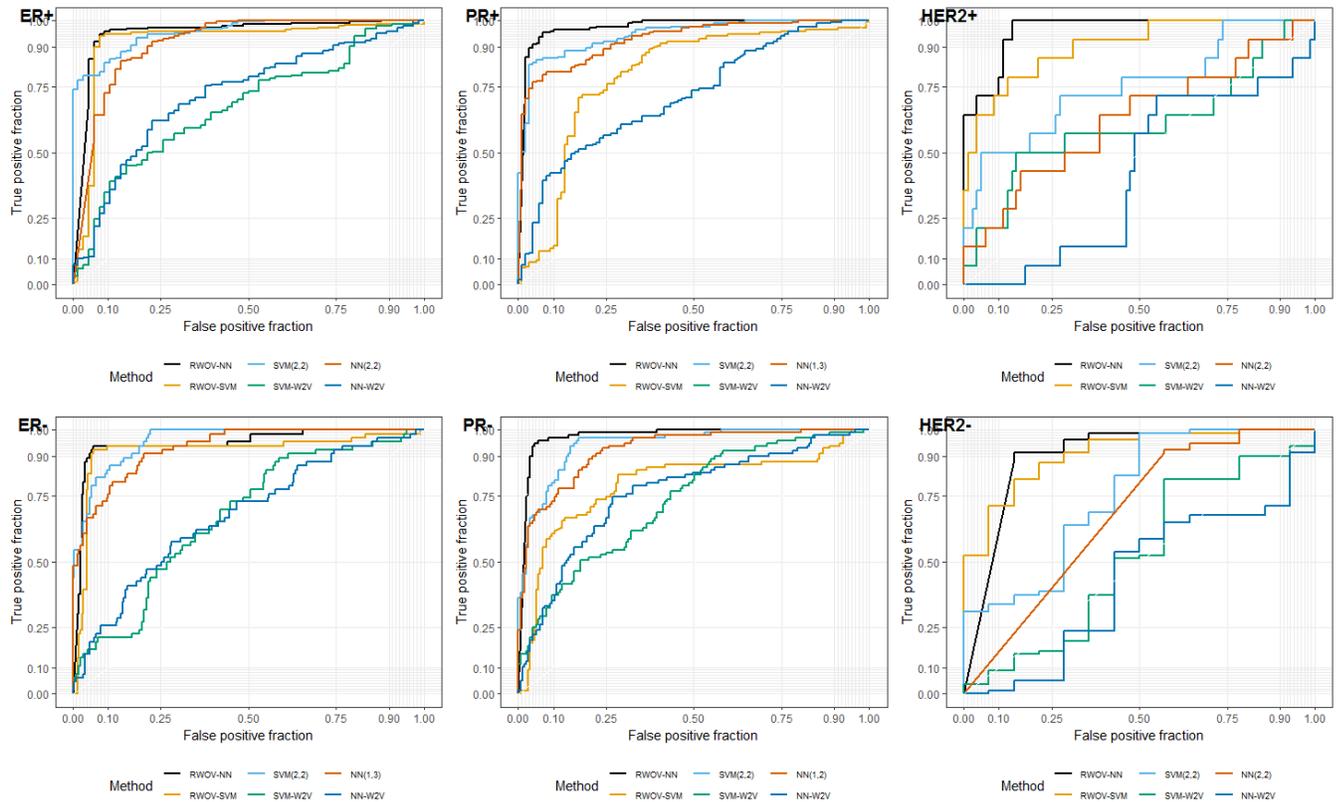

From Figures 1 and 2, it can be seen that in every case that there is a significantly superior method, it is RWOV that has the best performance. Figure 3 demonstrates that RWOV-NN typically exhibits the best cut threshold allowing for a high true positive fraction and a low false positive fraction.

## Discussion and Conclusions

Electronic healthcare records have enormous promise in facilitating research into improving patient treatment. However, much of the EHR is stored as unstructured data. Therefore, it is time-consuming to extract data from the EHR to pre-screen patients for clinical trials, or perform feasibility analysis for recruitment, because these records must be manually examined. Additionally, useful observational studies could be performed, if it were not for this major limitation. Nevertheless, given that the primary purpose of EHR is to support patient care, it would be inappropriate to change its structure to facilitate research. Therefore, it is imperative to develop methods for structuring and learning from these data that can facilitate these goals.

In this study, we have demonstrated that our method, Relevant Word Order Vectorization (RWOV), combined with a neural network, shows great promise in tackling this challenge. On a relevant use case, of identifying the hormone receptor status of breast cancer patients, RWOV showed consistently high accuracy across all three classification tasks. In most cases, it had the highest accuracy of any method examined in this study. Of particular importance, RWOV maintained high accuracy in classes with the poorest representation. This is necessary, because for some studies, it will be necessary to include patients based on these poorly represented classes, and poor accuracy might lead to some subjects being unnecessarily excluded (for example HER2+).

The reason for RWOV's performance on these tasks seems clear. It depends on a unique vectorization method that determines the most important words for classifying a particular case, in addition to their relative location. This relevant word ordering is well suited to the natural language processing in electronic health records, where the data are semi-structured, due to the repetitive nature of how healthcare providers often enter text. Our algorithm is able to take advantage of this semi-structured data to be a more powerful learner. The relatively poor performance of word2vec, which is a well-respected approach, is likely due to the small sample size. Typically, it depends on larger samples to perform well.

This initial approach, although promising, is only the beginning. There are many ways that our method could likely be improved. The learning algorithm was off the shelf, in order to provide proof-of-concept but would likely benefit from more customization. Also, we were limited in the amount of data that we could provide, due to the laborious process of hand-labeling examples that is required. Therefore, we think we could improve performance by implementing a semi-supervised approach, in addition to a larger training dataset.